\def\year{2020}\relax
\documentclass[letterpaper]{article} % DO NOT CHANGE THIS
\usepackage{aaai20}  % DO NOT CHANGE THIS
\usepackage{times}  % DO NOT CHANGE THIS
\usepackage{helvet} % DO NOT CHANGE THIS
\usepackage{courier}  % DO NOT CHANGE THIS
\usepackage[hyphens]{url}  % DO NOT CHANGE THIS
\usepackage{graphicx} % DO NOT CHANGE THIS
\usepackage{graphicx}  % DO NOT CHANGE THIS
\usepackage{amsmath}
\usepackage{amssymb}
\usepackage{multirow}
\usepackage{booktabs}

\pdfoutput=1

\title{Adaptive Unimodal Cost Volume Filtering for Deep Stereo Matching}
\author{
Youmin Zhang\textsuperscript{\rm 1}\footnotemark[\value{footnote}]\thanks{Joint first authorship. \{youmi,minwellcym\}@buaa.edu.cn.}
Yimin Chen\textsuperscript{\rm 1}\footnotemark[\value{footnote}]
Xiao Bai\textsuperscript{\rm 1}\thanks{Corresponding author, baixiao@buaa.edu.cn.}
Suihanjin Yu\textsuperscript{\rm 1}
Kun Yu\textsuperscript{\rm 2}
Zhiwei Li\textsuperscript{\rm 2}
Kuiyuan Yang\textsuperscript{\rm 2}
\\
\textsuperscript{\rm 1}State Key Laboratory of Software Development Environment, School of Computer Science and Engineering, \\ 
Beijing Advanced Innovation Center for Big Data and Brain Computing, \\
Jiangxi Research Institute, Beihang University, Beijing, China
\textsuperscript{\rm 2}DeepMotion
\\
\{youmi, minwellcym, baixiao, fakecoderemail\}@buaa.edu.cn
\{kunyu, zhiweili, kuiyuanyang\}@deepmotion.ai
}

\begin{document}
	
	\maketitle
	%------------------------------------------------------------------------
\begin{abstract}
State-of-the-art deep learning based stereo matching approaches treat disparity estimation as a regression problem, where loss function is directly defined on true disparities and their estimated ones. However, disparity is just a byproduct of a matching process modeled by cost volume, while indirectly learning cost volume driven by disparity regression is prone to overfitting since the cost volume is under constrained. In this paper, we propose to directly add constraints to the cost volume by filtering cost volume with unimodal distribution peaked at true disparities. In addition, variances of the unimodal distributions for each pixel are estimated to explicitly model matching uncertainty under different contexts. The proposed architecture achieves state-of-the-art performance on Scene Flow and two KITTI stereo benchmarks. In particular, our method
ranked the $1^{st}$ place of KITTI 2012 evaluation and the $4^{th}$ place of KITTI 2015 evaluation (recorded on 2019.8.20). The codes of AcfNet are available at: \url{https:// github.com/DeepMotionAIResearch/DenseMatchingBenchmark}.
\end{abstract}

\section{Introduction}
Stereo matching is one of the core technologies in computer vision, which recovers 3D structures of real world from 2D images. It has been widely used in areas such as autonomous driving~\cite{auto-driving}, augmented reality~\cite{augmented-reality} and robotics navigation~\cite{robot,luo2017vision,luo2019benchmark}. Given a pair of rectified stereo images, the goal of stereo matching is to compute the disparity $d$ for each pixel in the reference image (usually refers to the left image), where disparity is defined as the horizontal displacement between a pair of corresponding pixels in the left and right images. 

According to the seminar work~\cite{4steps}, a stereo matching algorithm typically consists of four steps: matching cost computation, cost aggregation, disparity regression and disparity refinement. Among them, matching cost computation, i.e., obtaining cost volume, is arguably the most crucial first step. A cost volume is usually denoted by a $H\times W\times D$ tensor, where $H$, $W$, $D$ are the height, width and maximal disparity of the reference image. In traditional methods, cost volume is computed by a predefined cost function of manually designed image features, e.g., squared or absolute difference of image patches.

\begin{figure}[t]
	\centering
	\includegraphics[width=0.95\columnwidth]{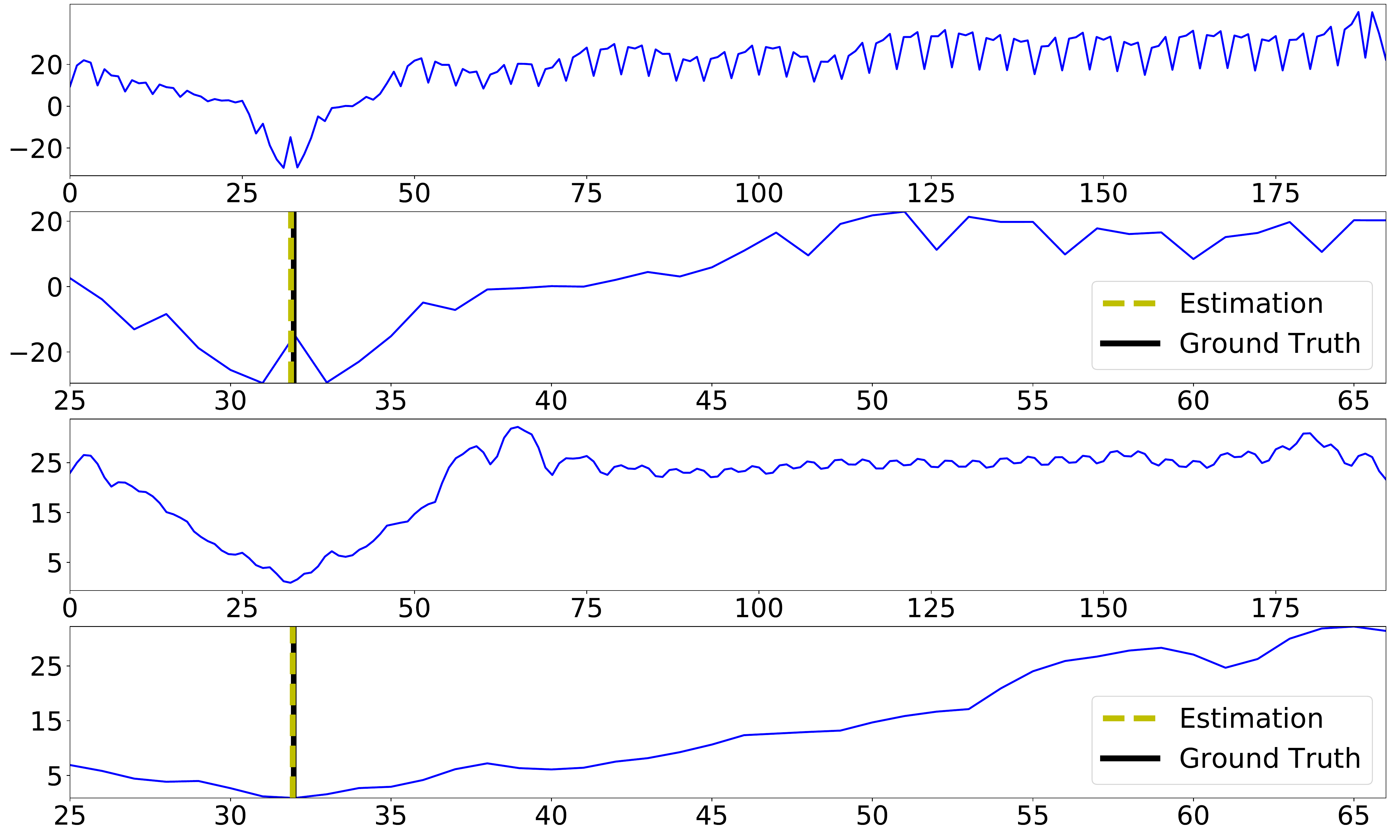}
	\caption{An example of cost distribution along the disparity dimension of cost volume. The first row is the output of PSMNet~\cite{PSM} trained with \textit{soft argmin}. The third row is output of our model. For better visualization, we also zoom into the disparity interval $[25, 66]$ in the second and fourth row, where Estimation and Ground Truth are the estimated and the groundtruth disparity respectively. Our method generates a more reasonable cost distribution peaked at the true disparity.}
	\label{fig:softargmin analysis}
\end{figure}

In the deep learning era, both image feature and cost function are modeled as network layers~\cite{GC-Net,PSM}. To make all layers differentiable and achieve sub-pixel estimation of disparity, \emph{soft argmin} is used to estimate disparity by softly weighting indices according to their costs, which is in contrast to \emph{argmin} that takes the index with minimal cost as estimated disparity. The loss function is defined on the estimated disparity and the ground truth for end-to-end training. Benefit from large-scale training data and end-to-end training, deep learning based stereo approaches achieve state-of-the-art performance.

\begin{figure*}
	\centering
		\includegraphics[width=2.0\columnwidth]{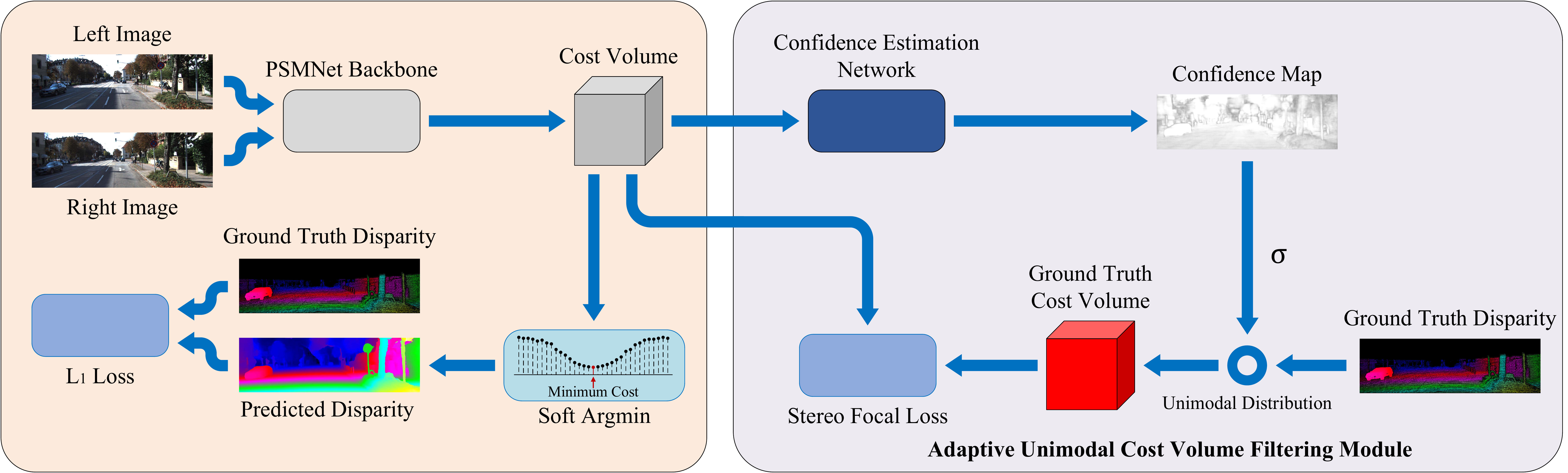}
		\caption{Architecture of the proposed end-to-end AcfNet. The input stereo images are fed to PSMNet~\cite{PSM} backbone with stacked hourglass architecture to get three cost volumes after aggregation. For each cost volume, we generate the confidence map by a Confidence Estimation Network (CENet), and modulate the ground truth cost volume with confidence values to generate pixel-wise unimodal distribution as training labels. The proposed \emph{Stereo Focal Loss} is added to the cost volume using the generated training labels. Finally, a sub-pixel disparity map is estimated by the \textit{soft argmin} function followed by regression loss as PSMNet.}
	\label{fig:framework}
\end{figure*}

In the deep learning models, the cost volume is indirectly supervised as an intermediate layer, which leaves cost volume less constrained since infinitely many cost distributions can generate the same disparity, where only cost distributions peaked at the true disparity are reasonable ones. Accordingly, we propose to directly supervise cost volume with unimodal ground truth distributions. To reveal network matching uncertainties~\cite{kendall2017uncertainties,ilg2018uncertainty} of different pixels, we design a confidence estimation network to estimate per-pixel confidence and control sharpness of the unimodal ground truth distributions accordingly. Figure~\ref{fig:softargmin analysis} compares the cost distributions at the same pixel by PSMNet and our method, where our method generates the correct minimal cost around the true disparity, while PSMNet generates two local minimal costs away from the true disparity.

We evaluate the proposed \textbf{A}daptive unimodal \textbf{c}ost volume \textbf{f}iltering \textbf{Net}work (AcfNet) on three stereo benchmarks including Scene Flow, KITTI 2012 and KITTI 2015. Ablation studies and detailed analysis on Scene Flow demonstrate the effectiveness of AcfNet. We also submit our stereo matching results to KITTI 2012 and 2015 evaluation server, and ranked the $1^{st}$ place on KITTI 2012 evaluation and the $4^{th}$ place on KITTI 2015 evaluation (recorded on 2019.8.20).

\section{Related Work}
Deep learning for stereo matching starts from learning image features for classical methods~\cite{mc-cnn,efficient}. DispNetC~\cite{sceneflow} is the first breakthrough for stereo matching by proposing an end-to-end trainable network, where cost function is predefined as a correlation layer in the network to generate the cost volume, then a set of convolutional layers are added to the cost volume to regress disparity map. Based on DispNetC, stack refinement sub-networks are proposed to improve the performance~\cite{CRL,IResNet}, and the performance could be further improved by using additional information such edges~\cite{EdgeStereo} and semantics~\cite{SegStereo}. To add more capacity for network to learn the cost function, \cite{GwcNet-gc} propose to use group-wise correlation layer and generate multiple cost volumes for latter aggregation. 

GC-Net~\cite{GC-Net} gives more flexibility for network to learn cost function by using 3D convolutional layers on concatenated feature volume, with cost volume produced by the learned cost function, disparity is estimated by \emph{soft argmin} according to the cost distribution. Follow-up works improve results by using better image features~\cite{PSM} and cost aggregation layers inspired by classical methods~\cite{cspn,GANet}. In these end-to-end stereo matching networks, cost volume is the output of an intermediate layer without direct supervision, which leaves the possibilities to learn unreasonable cost distributions as illustrated in Figure~\ref{fig:softargmin analysis}. 

In this work, the proposed AcfNet directly adds supervision to the cost volume estimation using ground truth cost distributions peaked at true disparities. In addition, the sharpness of ground truth cost distribution is adjusted according to matching confidence. Concurrent to our work, sparse LiDAR points are used to enhance cost volume by weighting estimated cost distribution with a Gaussian distribution centered at the disparity provided by its corresponding LiDAR point~\cite{GSM}, which serves as a multi-sensor fusion method for disparity estimation. In contrast, our method only takes images as input during both training and testing, and unimodal supervision is added to each pixel in a dense and adaptive way.

\section{AcfNet}
Figure~\ref{fig:framework} illustrates the overall framework, where the proposed adaptive unimodal cost volume filtering module is applied to the cost volume, and an additional loss is introduced to directly supervise the learning of cost volume towards desired property. Here, we choose PSMNet~\cite{PSM} as the basic network to calculate cost volume for its state-of-the-art performance on stereo matching.

\subsection{Overview}
Given a pair of rectified images, for each pixel $p=(x,y)$ in the left image, stereo matching aims to find its corresponding pixel in the right image, i.e., $p'=(x+d, y), d \in \mathbb{R}^+$, where disparity $d$ is often represented by a floating-point number for sub-pixel matching. For both computation and memory tractable, disparity is discreted into a set of possible disparities, i.e., $\{0, 1, \cdots, D-1\}$ to build an $H\times W \times D$ cost volume, where $H$, $W$ and $D$ are the image height, width and maximum disparity respectively. To recover sub-pixel matching, costs over disparities are used in a weighted interpolation. The whole process is implemented through a network as illustrated in the left part of Figure~\ref{fig:framework}.

Formally, the cost volume contains $D$ costs for each pixel denoted by $\{ c_{0}, c_{1}, \cdots, c_{D-1} \}$, and the sub-pixel disparity is estimated through \emph{soft argmin}~\cite{GC-Net}
\begin{equation}\label{soft-argmin}
\begin{aligned}
\hat{d} & = \sum_{d=0}^{D-1}d\times \hat{P}(d), 
\end{aligned}
\end{equation}
where $\hat{P}(d)=\frac{\exp{(-c_{d})}}{\sum_{d^{'}=0}^{D-1}\exp{(-c_{d^{'}})}}$, and disparities with small cost contribute more during interpolation. Given the groundtruth disparity $d_{p}$ for each pixel $p$, smooth $L_1$ loss is defined for training, i.e.,
\begin{equation}
    \begin{aligned}
    \mathcal{L}_{regression} = \frac{1}{|\mathcal{P}|}\sum_{p \in \mathcal{P}} smooth_{L_{1}}(d_{p} - \hat{d_{p}}),
    \end{aligned}
\end{equation}
where
\begin{equation}
    \begin{aligned}
    smooth_{L_{1}}(x) = 
    \Big\{
    \begin{array}{lc}
          0.5x^{2}, & \textrm{if}\; |x|<1,  \\
          |x| - 0.5, & \textrm{otherwise}.
    \end{array}
    \end{aligned}
\end{equation}
The whole process is differentiable by supervising with the groundtruth disparity, while cost volume is indirectly supervised through providing weights for disparity interpolation. However, the supervision is underdetermined and there could be infinitely possible sets of weights to achieve correct interpolation results. The flexibility of cost volume is prone to overfitting since many improperly learned cost volumes could interpolate disparities close to ground truth (i.e., small training loss).

To address this problem raised from indirectly supervising cost volume with underdetermined loss function, we propose to directly supervise the cost volume according to its unimodal property. 

\subsection{Unimodal distribution}
Cost volume is defined to reflect the similarities between candidate matching pixel pairs, where the true matched pair should have the lowest cost (i.e., the highest similarity), and the costs should increase with the distance to the truly matched pixel. This property requires unimodal distribution be peaked at the true disparity at each position in the cost volume. Given the ground truth disparity $d^{gt}$, the unimodal distribution is defined as
\begin{equation}\label{ground-truth similarity probability}
    \begin{aligned}
    P(d) & = \text{softmax}(-\frac{|d-d^{gt}|}{\sigma})
    \\  & = \frac{\exp{(-c_{d}^{gt})}}{\sum_{d^{'}=0}^{D-1}\exp{(-c_{d^{'}}^{gt})}},
    \end{aligned}
\end{equation}
where $c_{d}^{gt} = \frac{|d-d^{gt}|}{\sigma}$, $\sigma>0$ is the variance (a.k.a temperature in literature) that controls the sharpness of the peak around the true disparity.

The ground truth cost volume constructed from $P(d)$ has the same sharpness of peaks across different pixels, which cannot reflect similarity distribution differences across different pixels. For example, a pixel on the table corner should have a very sharp peak while pixels in uniform regions should have relative flat peaks. To build such more reasonable labels for cost volume, we add a confidence estimation network to adaptively predict $\sigma_p$ for each pixel.

\subsection{Confidence estimation network}
Considering matching properties are embedded in the estimated cost volume~\cite{fu,selecting,unified}, then the confidence estimation network takes the estimated cost volume as input, and uses a few layers to determine the matching confidence of each pixel by checking the matching states in a small neighborhood around each pixel. Specifically, the network employs a $3 \times 3$ convolutional layer followed by batch normalization and ReLU activation, and another $1\times 1$ convolutional layer followed by sigmoid activation to produce a confidence map $f \in [0,1]^{H\times W}$, where a pixel $p$ with large confidence $f_p$ means a unique matching can be confidently found for this pixel, while small confidence values denote there are matching ambiguities. Then, $\sigma_p$ for generating ground truth cost distribution is scaled from the estimated confidence,
\begin{equation}\label{sig}
    \begin{aligned}
    \sigma_{p} = s(1-f_{p})+ \epsilon,
    \end{aligned}
\end{equation}
where $s \geq 0$  is a scale factor that reflects the sensitivity of $\sigma$ to the change of confidence $f_p$, $\epsilon>0$ defines the lower bound for $\sigma$ and avoids numerical issue of dividing 0. Accordingly, $\sigma_p \in [\epsilon, s + \epsilon]$. Our experiments show that two kinds of pixels are likely to have large $\sigma$, i.e., texture-less pixels and occluded pixels, where the texture-less pixels tend to have multiple matches, while occluded pixels have no correct matches. With the per-pixel adpatively estimated $\sigma_p$, the ground truth cost volume defined in Eq.~(\ref{ground-truth similarity probability}) is modified accordingly.

\subsection{Stereo focal loss}
At pixel position $p$, we now have both estimated cost distribution $\hat{P}_p(d)$ and the ground truth $P_p(d)$. It is straightforward to define a distribution loss via cross entropy. However, there is a severe sample imbalance problem since each pixel has only one true disparity (positive) comparing with hundreds of negative ones~\cite{mc-cnn}. Similar to focal loss designed to solve the sample imbalance problem in one-stage object detection ~\cite{focalLoss}, we design a stereo focal loss to focus on positive disparities to avoid the total loss dominated by negative disparities, 
\begin{equation}\label{SF}
%\footnotesize
\scriptsize
\begin{aligned}
\mathcal{L}_{SF} = \frac{1}{|\mathcal{P}|}\sum_{p \in \mathcal{P}}\left(\sum_{d=0}^{D-1}(1 - P_{p}(d))^{-\alpha} \cdot \left( - P_{p}(d) \cdot \log \hat{P}_{p}(d) \right) \right),
\end{aligned}
\end{equation}
where $\alpha \ge 0$ is a \emph{focusing} parameter, and the loss is reduced to cross entropy loss when $\alpha=0$, while $\alpha > 0$ gives more weights to positive disparities in proportion to their $P_p(d)$.
Thus easy negative disparities are further suppressed explicitly with quite small weights and let the positive disparity only compete with a few hard ones.

\subsection{Total loss function}
In sum, our final loss function contains three parts defined as
\begin{equation}\label{all loss}
\begin{aligned}
\mathcal{L} = \mathcal{L}_{SF} & + \lambda_{regression} \mathcal{L}_{regression} \\ & + \lambda_{confidence} \mathcal{L}_{confidence},
\end{aligned}
\end{equation}
where $\lambda_{regression}, \lambda_{confidence}$ are two trade-off hyper-parameters. $\mathcal{L}_{SF}$ supervises the cost volume while $\mathcal{L}_{regression}$ supervises the disparity. $\mathcal{L}_{confidence}$ is added as a regularizer to encourage more pixels to have high confidence values,
\begin{equation}\label{CEN}
\begin{aligned}
\mathcal{L}_{confidence} = \frac{1}{|\mathcal{P}|}\sum_{p \in \mathcal{P}}-\log f_{p}.
\end{aligned}
\end{equation}

\section{Experiments and Analysis}

\subsection{Implementation details}
Our network is implemented using PyTorch~\cite{pytorch} framework, and all models are end-to-end trained using RMSprop with standard settings. Our data processing is the same as PSMNet~\cite{PSM}. We train our models from scratch using the Scene Flow dataset with a constant learning rate of 0.001 for 10 epochs. For Scene Flow, the trained model is directly used for testing. For KITTI, we use the model trained with Scene Flow data after fine-tuning on the KITTI training set for 600 epochs. The learning rate of this fine-tuning begins at 0.001 and is decayed by $\frac{1}{3}$ at 100 and 300 epochs. For submission to the KITTI test benchmark, we prolong the training process on Scene Flow with a constant learning rate of 0.001 for 20 epochs to obtain a better pre-training model. The batch size is set to 3 for training on 3 NVIDIA GTX 1080Ti GPUs. All ground truth disparities out of range of $[0, D-1]$ are excluded in our experiments, where $D=192$.

\subsection{Datasets}
We evaluate AcfNet qualitatively and quantitatively on three challenging stereo benchmarks, i.e., Scene Flow~\cite{sceneflow}, KITTI 2012~\cite{KITTI2012} and KITTI 2015~\cite{KITTI2015}. 

\noindent
\textbf{Scene Flow:} Scene Flow is a large synthetic dataset containing 35,454 training image pairs and 4,370 testing image pairs, where the ground truth disparity maps are densely provided, which is large enough for directly training deep learning models. Following the setting as GC-Net~\cite{GC-Net}, we mainly use this dataset for ablation study.

\noindent
\textbf{KITTI:} KITTI 2015 and KITTI 2012 are two real-world datasets with street views captured from a driving car. KITTI 2015 contains 200 training stereo image pairs with sparse groundtruth disparities obtained using LiDAR and 200 testing image pairs with ground truth disparities held by evaluation server for submission evaluation only. KITTI 2012 contains 194 training image pairs with sparse ground truth disparities and 195 testing image pairs with ground truth disparities held by evaluation server for submission evaluation only. These two datasets are challenging due their small size. 

\noindent
\textbf{Metrics:} The performance is measured using two standard metrics: (1) {3-Pixel-Error} (3PE), i.e., the percentage of pixels for which the predicted disparity is off the true one by more than 3 pixels, and (2) {End-Point-Error} (EPE), i.e., the average difference of the predicted disparities and their true ones. 3PE is robust to outliers with large disparity errors, while EPE measures errors to sub-pixel level.

To further evaluate the ability on handling occluded regions, we divide the testing images of Scene Flow into occluded region (OCC) and non-occluded regions (NOC) through left-right consistency check. In total, there are 16\% occluded pixels in all pixels. The performance is measured on all pixels if no prefix such as OCC, NOC and ALL are added before 3PE or EPE.

\begin{figure*}[!h]
	\begin{center}
		\includegraphics[width=2.0\columnwidth]{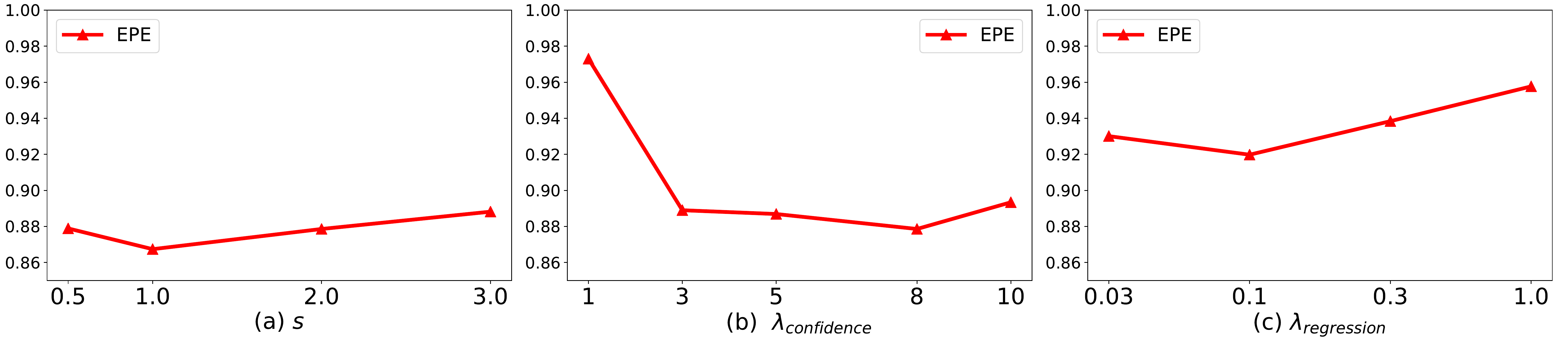}
	\end{center}
	\caption{Ablation study results for different hyper-parameters in our method, where $s$ controls the upper bound of variance $\sigma$. $\lambda_{confidence}$ and $\lambda_{regression}$ are balance weights for confidence loss and disparity regression loss respectively.}
	\label{fig:parameter}
\end{figure*}
\begin{figure}[!h]
	\begin{center}
		\includegraphics[width=0.95\columnwidth]{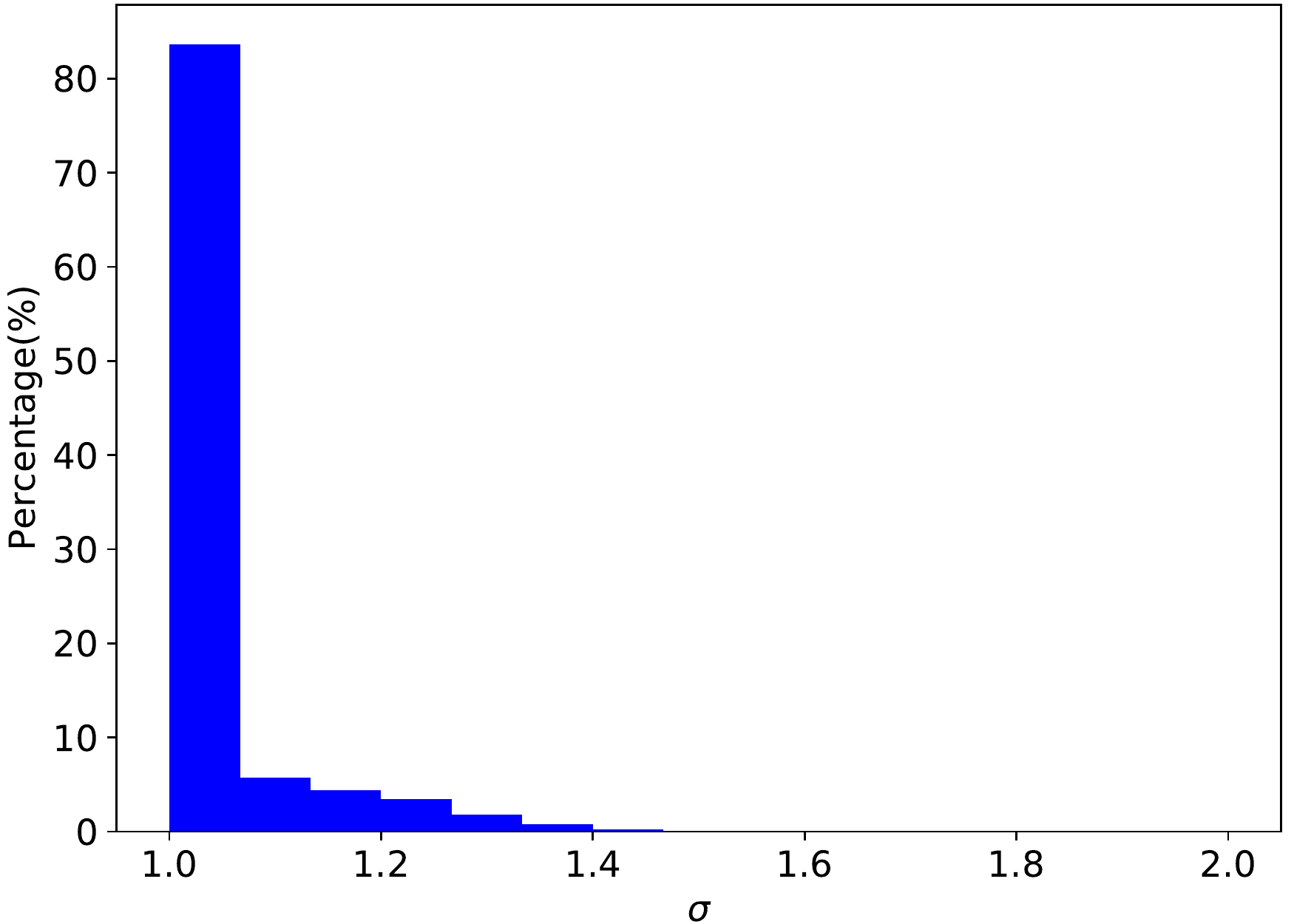}
	\end{center}
	\caption{Histogram distribution of variance $\sigma$ on the whole test dataset of Scene Flow after AcfNet has been converged. }
	\label{fig:sigma}
\end{figure}

\subsection{Ablation studies}
We conduct ablation studies on Scene Flow~\cite{sceneflow} considering it has large enough training data for end-to-end training from scratch. In all experiments, $\alpha$ is set to 5.0 in stereo focal loss to balance positive and negative samples. Considering disparities of most pixels are with sub-pixel errors (i.e., error smaller than one pixel) while 3PE cannot reveal errors within 3 pixels, we use EPE to study the performance variance for different hyper-parameter settings.

\begin{table}[!h]
	\caption{Results of comparison between stereo focal loss and cross entropy loss in our model AcfNet.}
	\begin{center}
		\resizebox{0.95\columnwidth}{!}{
			\begin{tabular}{l*{2}{c}}
				\toprule
				 AcfNet & + Cross Entropy Loss & + Stereo Focal Loss \\
				 
				 \midrule
				 
				 EPE [px] & 0.965 & \textbf{0.920} \\
				
				\bottomrule	
			\end{tabular}
		}
	\end{center}
	\label{sf comparation}
\end{table}

\noindent
{\bf The variance $\sigma$ of unimodal distribution}

The variance $\sigma$ adjusts the shape of unimodal distribution, which plays an important role in AcfNet. In our method, $\sigma \in [\epsilon, s + \epsilon]$ is bounded by $s$ and $\epsilon$.

Firstly, we study the case when the variance $\sigma$ is fixed for all pixels, i.e. $s=0, \sigma=\epsilon$. By grid search, we find that $\sigma=1.2$ achieves the best result, which indicates most pixels favor $\sigma=1.2$ for building unimodal distributions. Thus, we set the lower bound $\epsilon$ of $\sigma$ to 1.0 for adaptive variance study. Furthermore, we compare the stereo focal loss with cross entropy loss under this condition, i.e. $\sigma=1.2$. As shown in Table~\ref{sf comparation}, equipping AcfNet with stereo focal loss get a significantly better result than cross entropy loss, which  demonstrates the effectiveness of stereo focal loss in balancing losses from positive and negative disparities.

Secondly, we study the sensitivity $s$ which controls the upper bound of $\sigma$. Figure~\ref{fig:parameter}(a) shows the performance by varying $s$, where $s=1$ performs best and the performance is rather stable by varying $s$ from 0.5 to 3.0. Figure~\ref{fig:sigma} shows the histogram of $\sigma$ when $s=1$ (i.e., $\sigma \in [1.0, 2.0]$), where most pixels favor small variances, i.e., sharp distributions, and a long tail of pixels require larger variances for flatten distributions.

\noindent
\textbf{Loss balance weights}

Hyperparameter $\lambda_{confidence}$ balances the total variance and other losses. Figure.~\ref{fig:parameter}(b) shows the performance curve by varying $\lambda_{confidence}$, where both overconfident learning with large $\lambda_{confidence}$ and underconfident learning with small $\lambda_{confidence}$ lead to inferior performance while $\lambda_{confidence}=8.0$ performs the best.

Hyperparameter $\lambda_{regression}$ balances the regression loss that is widely used in recent state-of-the-art models, and large value for $\lambda_{regression}$ will eliminate effects of the other two losses proposed in this paper. Figure~\ref{fig:parameter}(c) shows the performance curve, it could be observed that regression loss can be improved through proper tradeoff with the proposed two losses.

\begin{figure}[!h]
	\centering
	\includegraphics[width=0.95\columnwidth]{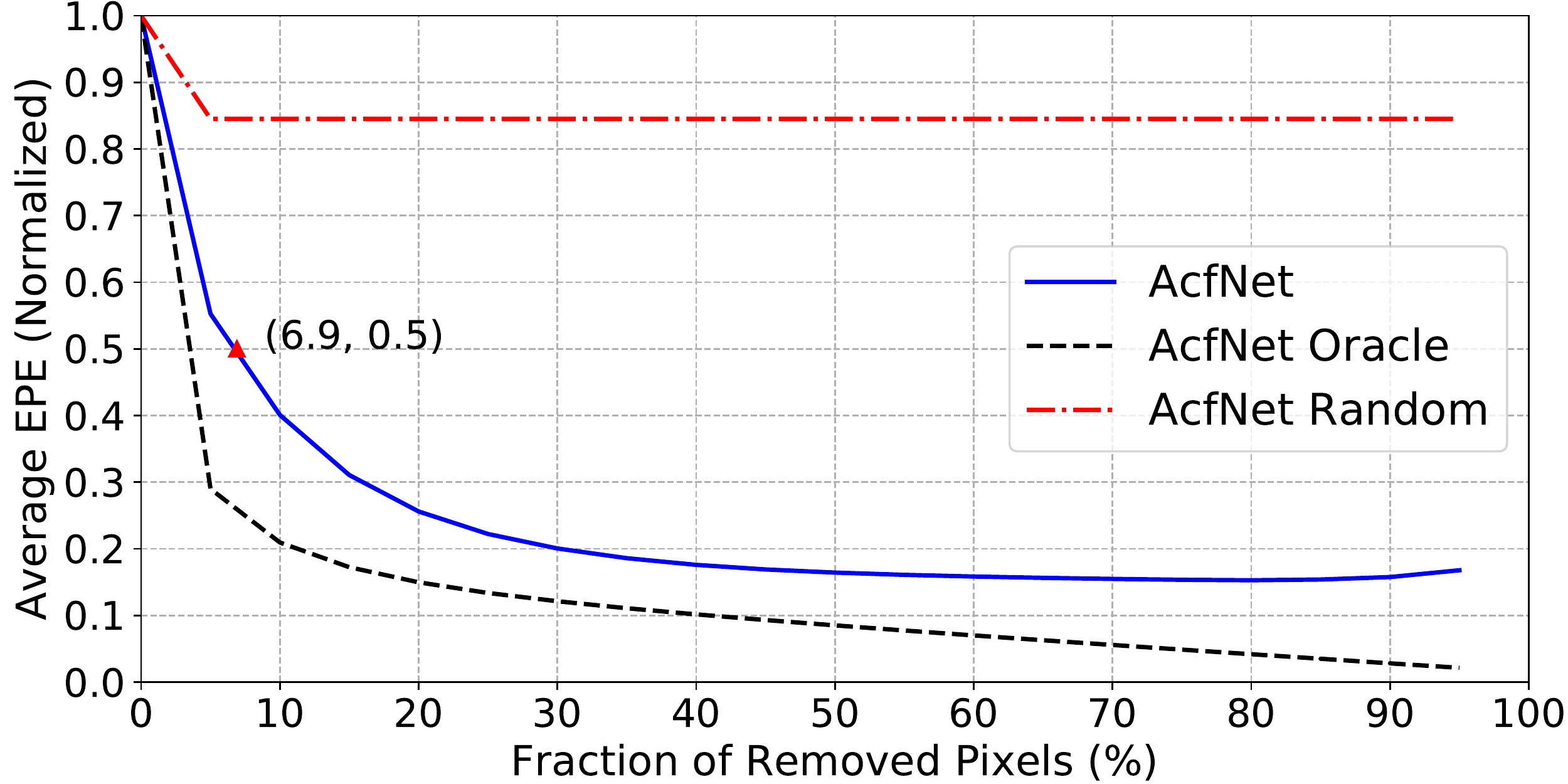}
	\caption{Sparsification plot of our AcfNet on the Scene Flow test dataset. The plot shows the normalized average end-point-error (EPE) for each fraction of pixels with highest variances has been removed. The curve `AcfNet Oracle' shows the ideal case by removing each fraction of pixels ranked by the ground truth EPE. The curve `AcfNet Random' shows the worst case by removing each fraction of pixels randomly. Removing only 6.9\% of the pixels by AcfNet results in  halving the average EPE. }
	\label{fig:sparsification}
\end{figure}

\begin{figure*}[!h]
	\begin{center}
		\includegraphics[width=2.0\columnwidth]{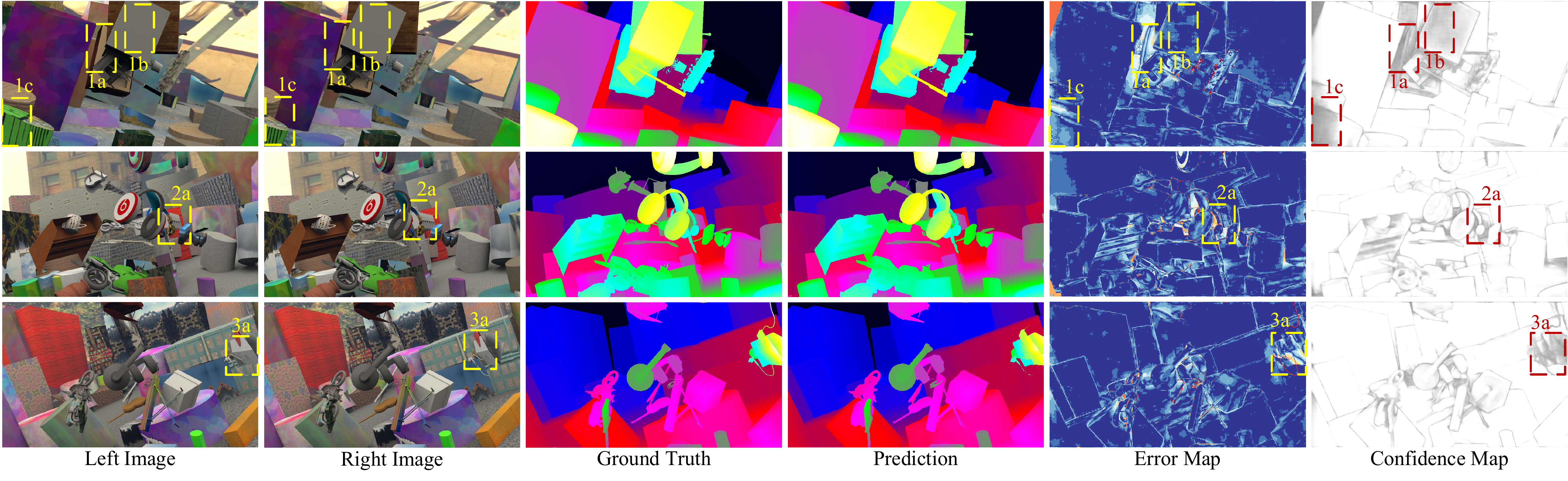}
	\end{center}
	\caption{Qualitative results on three samples from Scene Flow test set. Columns from left to right are: left stereo input image, right stereo input image, disparity ground truth, disparity prediction, error map and confidence map. Cold colors in the error map denote small prediction errors while warm colors denote large prediction errors. In confidence map, bright colors mean small variances while dark colors denote high variances.}
	\label{fig:sceneflow-example}
\end{figure*}

\subsection{Variance analysis}
Variance estimation is an important component of our cost filtering scheme, which automatically adjusts the flatness of the unimodal distribution according to the matching uncertainty. To assess the quality of the estimated variances, sparsification plot~\cite{ilg2018uncertainty} is adopted to reveal the relevance of the estimated variances with the true errors through plotting evaluation results by gradually removing pixels according their variances. For comparison, we also plot the curves of randomly assigned variances (AcfNet Random) and variances assigned by EPE errors (AcfNet Oracle) in Figure~\ref{fig:sparsification}, where the estimated variances are highly relevant to EPE errors and demonstrates the ability of AcfNet in explaining outlier pixels with estimated variances.

Figure~\ref{fig:sceneflow-example} shows several per-pixel results from Scene Flow, where hard regions mainly appear at occlusions (1a, 1c and 2a), repeated patterns (1b, 3a) and thin structures (3a). In these hard regions, AcfNet provides high variances to flatten the corresponding cost distributions. AcfNet can balance the learning for different pixels, and pushes informative pixels towards high confidences (i.e, low variances), while allows hard uninformative pixels with high variances to avoid overfitting.

\begin{table}[!htbp]
	\caption{Evaluation of adaptive unimodal cost volume filtering results, where PSMNet is re-implemented. AcfNet(uniform) denotes setting a uniform unimodal distribution for all pixels,  and AcfNet(adaptive) denotes adaptively adjust the per-pixel variances. }
	\begin{center}
        \resizebox{0.95\columnwidth}{!}{
			\begin{tabular}{l*{3}{c}*{3}{c}}
				\toprule
				\multirow{3}*{Method} &	\multicolumn{6}{c}{Scene Flow} \\
				\cline{2-7}
				& \multicolumn{3}{c}{EPE [px]} &
				\multicolumn{3}{c}{3PE [\%]} \\
				\cline{2-7}
				& ALL & OCC & NOC & ALL & OCC & NOC \\
				
				\midrule
				PSMNet  & 1.101 & 3.507 & 0.637 & 4.56 & 17.64 & 2.12 \\
				AcfNet (uniform) & 0.920 & 2.996 & 0.504 & 4.39 & 16.47 & \textbf{2.10} \\
				AcfNet (adaptive) &  \textbf{0.867} & \textbf{2.736} & \textbf{0.495} & \textbf{4.31} & \textbf{15.77} & 2.13 \\ 
				
				\bottomrule	
			\end{tabular}
		}
	\end{center}
	\label{network setting comparation}
\end{table}

\subsection{Adaptive unimodal cost volume filtering}
AcfNet adds direct cost volume supervision to PSMNet. Table~\ref{network setting comparation} compares two versions of AcfNet with PSMNet, where uniform version of AcfNet is significantly better than PSMNet and adaptive version of AcfNet further improves the performance significantly. The results demonstrate the effectiveness of unimodal supervision and adaptive per-pixel variance estimation. Comparing with AcfNet(uniform), AcfNet(adaptive) improves more on OCC (i.e., occluded regions), which is consistent with conclusion in variance analysis.

\begin{table}[!h]
	\caption{Results of cost volume filtering comparison, where all methods are trained on Scene Flow from scratch using the same base model PSMNet, and directly test on KITTI 2012, 2015 training datasets. $^{*}$ denotes disparities of sparse LiDAR points are also used as model input when testing.}
	\begin{center}
        \resizebox{0.95\columnwidth}{!}{
			\begin{tabular}{lc|*{2}{c}}
				\toprule
				\multirow{2}*{Method} & \multicolumn{1}{c}{EPE[px]} & \multicolumn{2}{c}{3PE[\%]} \\
				\cline{2-4}
				& Scene Flow& KITTI 2012 & KITTI 2015 \\
				\hline
				PSMNet & 1.101 & 29.18 & 30.19 \\
				\cite{GSM} & $0.991^{*}$ & - & $23.13^{*}$ \\
				AcfNet &  \textbf{0.867} & \textbf{17.54} & \textbf{19.45}  \\
				\bottomrule
			\end{tabular}
		}
	\end{center}
	\label{generalization}
\end{table}

\subsection{Cost volume filtering comparisons}
To further validate the superiority of the proposed cost volume filtering, experiments are designed to compare with the concurrent work~\cite{GSM}. In contrast to our work, \cite{GSM} uses disparities by sparse LiDAR points to filter cost volume during both training and testing. Both AcfNet and the method of \cite{GSM} are trained on Scene Flow from scratch, and directly evaluated on training sets of KITTI 2012 and 2015 since \cite{GSM} requires sparse LiDAR points as inputs. Table~\ref{generalization} reports the comparison results, where AcfNet outperforms \cite{GSM} on all performance metrics by large margins even without using LiDAR points as inputs. In addition, comparing with PSMNet, AcfNet shows much better generalization performance from Scene Flow to KITTI, which further proves the ability of AcfNet in preventing overfitting.

\subsection{Comparisons with the state-of-the-art methods}
\begin{figure*}[!h]
	\begin{center}
		\includegraphics[width=2.0\columnwidth]{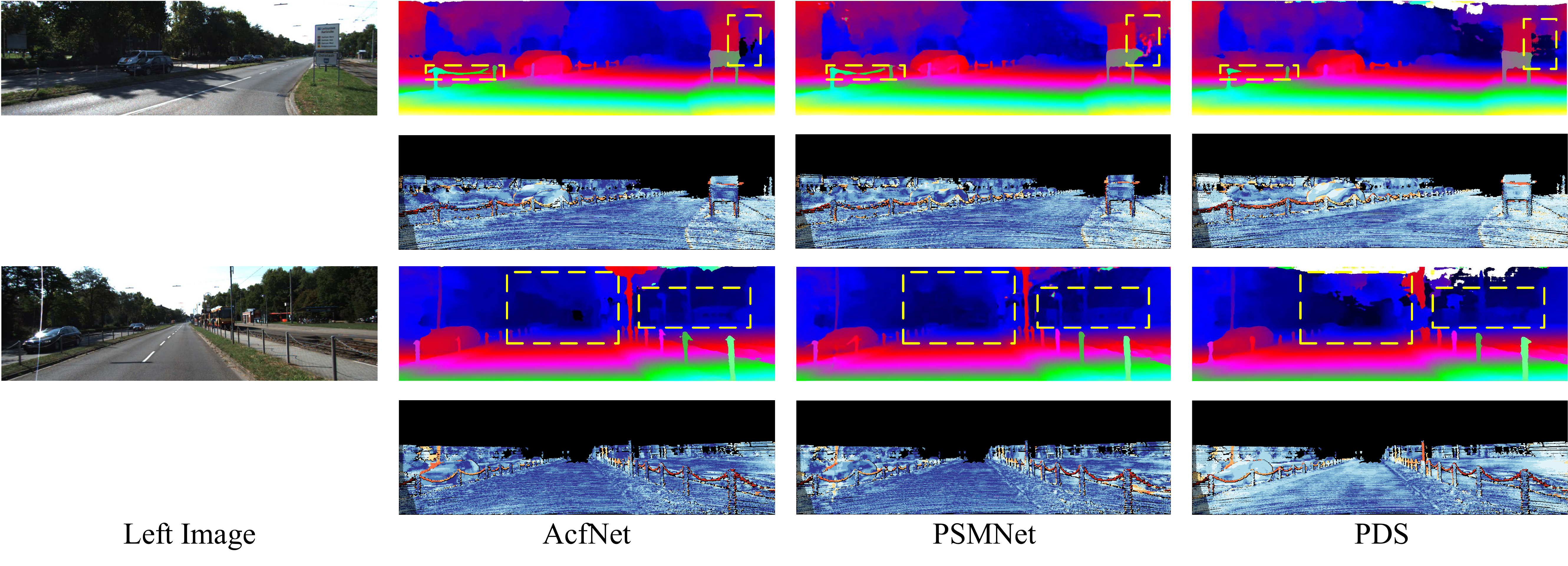}
	\end{center}
	\caption{Visualization results on the KITTI 2015 dataset. Significantly improved regions are highlighted with dash boxes. For each example, the first row shows the disparity map, and the second row shows the error map. Warmer color indicate larger prediction errors.}
	\label{fig:KITTI2015}
\end{figure*}

\begin{figure*}[!h]
	\begin{center}
		\includegraphics[width=2.0\columnwidth]{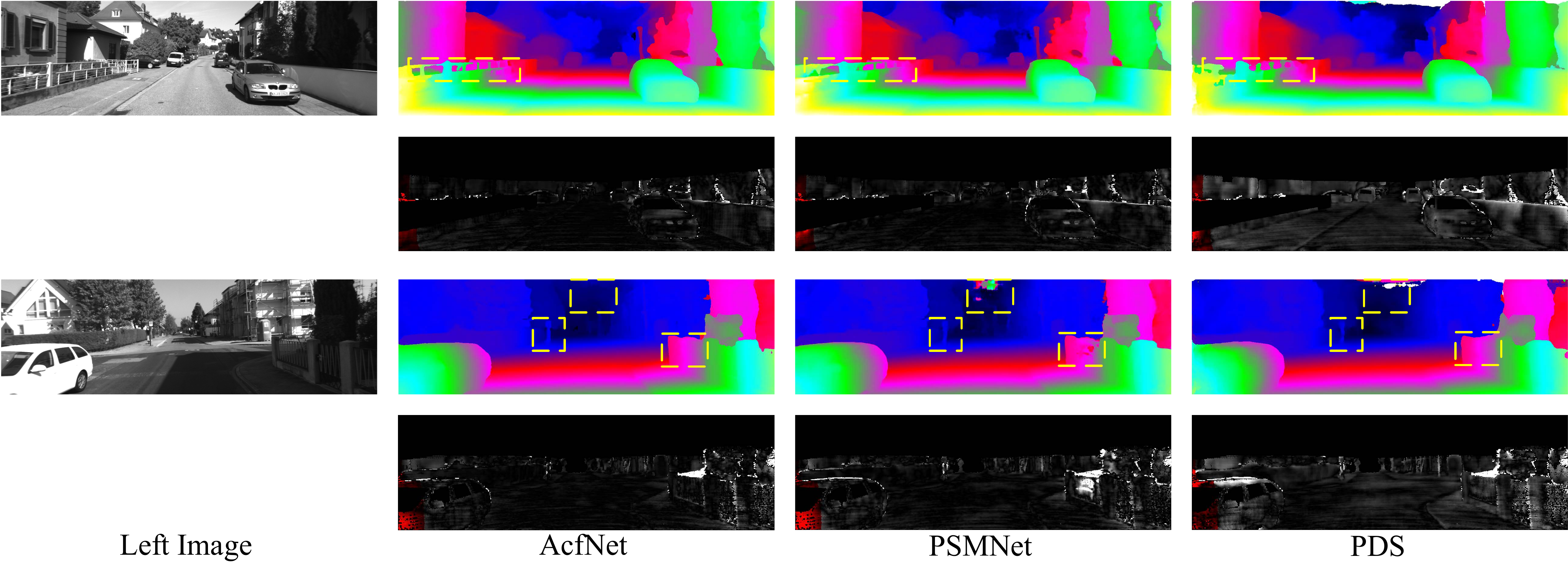}
	\end{center}
	\caption{Visualization results on the KITTI 2012 dataset. Significantly improved regions are highlighted with dash boxes. For each example, the first row shows the disparity map, and the second row shows the error map, bright colors indicate inaccurate predictions.}
	\label{fig:KITTI2012}
\end{figure*}

To further validate the proposed AcfNet, Table~\ref{benchmark} compares AcfNet with state-of-the-art methods on both KITTI 2012 and 2015, where AcfNet outperforms others by notable margins on all evaluation metrics. To be noted, Scene Flow is used for pretraining in all methods considering the small size of KITTI training data. Figure~\ref{fig:KITTI2015} and~\ref{fig:KITTI2012} show several exemplar results from KITTI 2015 and 2012 by comparing AcfNet with PSMNet~\cite{PSM} and PDS~\cite{pds}, where significantly improved regions are marked out with dash boxes. As expected, most improvements of AcfNet come from challenging areas such as thin structures, sky boundaries and image borders.

\begin{table*}[!h]
	\caption{Results on Scene Flow and KITTI Benchmarks. Following standard setting, on KITTI 2012, percentages of erroneous pixels for both Non-occluded (Out-Noc) and all (Out-All) pixels are reported, on KITTI 2015, percentages of disparity outliers $D_{1}$ averaged over all ground truth pixels (D1-all) for both Non-occluded and All pixels are reported. The outliers are defined as those pixels whose disparity errors are larger than $\max(3\textrm{px}, 0.05d^{gt})$, where $d^{gt}$ is the ground-truth disparity.}
	\begin{center}
        \resizebox{2.0\columnwidth}{!}{
			\begin{tabular}{lc|*{8}{c}|c c}
				\toprule
				\multirow{3}*{Method} & \multicolumn{1}{c}{Scene Flow} & \multicolumn{8}{c}{KITTI 2012} &\multicolumn{2}{c}{KITTI 2015} \\
				\cline{2-12}
				& \multirow{2}*{EPE} & \multicolumn{2}{c}{2px} & \multicolumn{2}{c}{3px} & \multicolumn{2}{c}{4px} & \multicolumn{2}{c}{5px} & \multicolumn{1}{|c}{ALL} & NOC \\
				& & Out-Noc & Out-All & Out-Noc & Out-All & Out-Noc & Out-All & Out-Noc & Out-All & D1-all & D1-all\\
				\hline
				MC-CNN~\cite{mc-cnn} & 3.79 & 3.90 & 5.45 & 2.43 & 3.63 & 1.90 & 2.85 & 1.64 & 2.39 & 3.88 & 3.33 \\
				GC-Net~\cite{GC-Net} & 2.51 & 2.71 & 3.46 & 1.77 & 2.30 & 1.36 & 1.77 & 1.12 & 1.46 & 2.67 & 2.45 \\
				iResNet-i2~\cite{IResNet} & 1.40 & 2.69 & 3.34 & 1.71 & 2.16 & 1.30 & 1.63 & 1.06 & 1.32 & 2.44 & 2.19 \\
				PSMNet~\cite{PSM} & 1.09 & 2.44 & 3.01 & 1.49 & 1.89 & 1.12 & 1.42 & 0.90 & 1.15 & 2.32 & 2.14 \\
				
				% EdgeStereo~\cite{EdgeStereo} & 1.12 & 2.79 & 3.43 & 1.73 & 2.18 & 1.30 & 1.64 & 1.04 & 1.32 & 2.16 & 2.00 \\
				
				SegStereo~\cite{SegStereo} & 1.45 & 2.66 & 3.19 & 1.68 & 2.03 & 1.25 & 1.52 & 1.00 & 1.21 & 2.25 & 2.08 \\
				
				PDS~\cite{pds} & 1.12 & 3.82 & 4.65 & 1.92 & 2.53 & 1.38 & 1.85 & 1.12 & 1.51 & 2.58 & 2.36 \\
				
				GwcNet-gc~\cite{GwcNet-gc} & \textbf{0.77} & 2.16 & 2.71 & 1.32 & 1.70 & 0.99 & 1.27 & 0.80 & 1.03 & 2.21 & 1.92 \\
				
				HD$^{3}$-Stereo~\cite{HD3} & 1.08 & 2.00 & 2.56 & 1.40 & 1.80 & 1.12 & 1.43 & 0.94 & 1.19 & 2.02 & 1.87 \\
				
				GA-Net~\cite{GANet}& 0.84 & 2.18 & 2.79 & 1.36 & 1.80 & 1.03 & 1.37 & 0.83 & 1.10 & 1.93 & 1.73 \\
				
				\hline
				
				AcfNet & 0.87 & \textbf{1.83} & \textbf{2.35} & \textbf{1.17} & \textbf{1.54} & \textbf{0.92} & \textbf{1.21} & \textbf{0.77} & \textbf{1.01} & \textbf{1.89} & \textbf{1.72} \\
				
				\bottomrule
			\end{tabular}
		}
	\end{center}
	\label{benchmark}
\end{table*}

\section{Conclusions}
In this paper, we solve the under-constrain problem of cost volume in existing deep learning based stereo matching approaches. The proposed AcfNet supervises the cost volume with ground truth unimodal distributions peaked at true disparities, and variances for per-pixel distributions are adaptively estimated to modulate the learning according the informativeness of each pixel. AcfNet shows better testing performance on the same dataset and even superior performance on cross-dataset evaluation.

\section{Acknowledgements}
This work was supported by the National Natural Science Foundation of China project no. 61772057 and the support funding from State Key Lab. of Software Development Envi- ronment and Jiangxi and Qingdao Research Institute of Bei- hang University.

\section{Appendices}

\subsection{A. Effectiveness on different backbones}

We evaluate the effectiveness of our adaptive unimodal cost volume filtering scheme among different backbones, namely, the stack-hourglass version of PSMNet~\cite{PSM} and GC-Net~\cite{GC-Net}. We re-implement all methods with the training protocol detailed in \textbf{Implementation details}. Specifically, the batch size of GC-Net is set to 24 for training on 8 Tesla V100.  Table~\ref{effOnModels} reports the results, our method delvers better performance across different backbones.

\begin{table}[!htbp]
	\caption{Evaluation our method among different stereo matching models, where * denotes equipping the model with our adaptive unimodal cost volume filtering scheme. }
	\begin{center}
        \resizebox{0.95\columnwidth}{!}{
			\begin{tabular}{l*{3}{c}*{3}{c}}
				\toprule
				\multirow{3}*{Method} &	\multicolumn{6}{c}{Scene Flow} \\
				\cline{2-7}
				& \multicolumn{3}{c}{EPE [px]} &
				\multicolumn{3}{c}{3PE [\%]} \\
				\cline{2-7}
				& ALL & OCC & NOC & ALL & OCC & NOC \\
				
				\midrule
				
				GC-Net & 0.871 & 2.916 & 0.452 & \textbf{3.89} & \textbf{15.63} & \textbf{1.65} \\
				GC-Net* & \textbf{0.822} & \textbf{2.777} & \textbf{0.436} & 4.33 & 16.46 & 2.02 \\
				
				\midrule

				PSMNet  & 1.101 & 3.507 & 0.637 & 4.56 & 17.64 & \textbf{2.12} \\
				PSMNet* &  \textbf{0.867} & \textbf{2.736} & \textbf{0.495} & \textbf{4.31} & \textbf{15.77} & 2.13 \\
				
				\bottomrule	
			\end{tabular}
		}
	\end{center}
	\label{effOnModels}

\end{table}

\subsection{B. Architecture details}

Table~\ref{network parameters} presents the details of the AcfNet which is used in experiments to produce state-of-the-art accuracy on Scene Flow dataset~\cite{sceneflow} and KITTI benchmarks~\cite{KITTI2012,KITTI2015}. It is based on PSMNet with stacked hourglass architecture, which produces three cost volumes, and Confidence Estimation network(CENet) is added to each of the cost volume.

\begin{table}[!htbp]
	\caption{Parameters of the network architecture of AcfNet.}
	\begin{center}
        \resizebox{0.95\columnwidth}{!}{
			\begin{tabular}{c|c|c}
				\toprule
				 Name & Layer setting & Output dimension \\

				 \midrule
				 
				 \multicolumn{3}{c}{\textbf{Feature Extraction}} \\
				 
				 \midrule
				 
				 input & & $H \times W \times 3$ \\
				 \hline
				 conv0\_x & $[3\times 3, 32]\times 3$ & $\frac{1}{2}H\times \frac{1}{2}W \times 32$ \\
				 \hline
				 conv1\_x & $\left[ \begin{array}{c} 3\times 3, 32 \\ 3\times 3, 32  \end{array} \right] \times 3 $ & $\frac{1}{2}H\times \frac{1}{2}W \times 32$ \\
				 \hline
				 conv2\_x & $\left[ \begin{array}{c} 3\times 3, 64 \\ 3\times 3, 64  \end{array} \right] \times 16 $ & $\frac{1}{4}H\times \frac{1}{4}W \times 64$ \\
				 \hline
				 conv3\_x & $\left[ \begin{array}{c} 3\times 3, 128 \\ 3\times 3, 128  \end{array} \right] \times 3, dila =2 $ & $\frac{1}{4}H\times \frac{1}{4}W \times 128$ \\
				 \hline
				 conv4\_x & $\left[ \begin{array}{c} 3\times 3, 128 \\ 3\times 3, 128  \end{array} \right] \times 3, dila =4 $ & $\frac{1}{4}H\times \frac{1}{4}W \times 128$ \\
				 \hline
				 branch\_1 & $ \begin{array}{c} 64\times 64, \textrm{avg.pool} \\ 3\times 3, 32 \\ \textrm{bilinear interpolation}  \end{array} $ & $\frac{1}{4}H\times \frac{1}{4}W \times 32$ \\
				 \hline
				 branch\_2 & $ \begin{array}{c} 32\times 32, \textrm{avg.pool} \\ 3\times 3, 32 \\ \textrm{bilinear interpolation}  \end{array} $ & $\frac{1}{4}H\times \frac{1}{4}W \times 32$ \\
				 \hline
				 branch\_3 & $ \begin{array}{c} 16\times 16, \textrm{avg.pool} \\ 3\times 3, 32 \\ \textrm{bilinear interpolation}  \end{array} $ & $\frac{1}{4}H\times \frac{1}{4}W \times 32$ \\
				 \hline
				 branch\_4 & $ \begin{array}{c} 8\times 8, \textrm{avg.pool} \\ 3\times 3, 32 \\ \textrm{bilinear interpolation}  \end{array} $ & $\frac{1}{4}H\times \frac{1}{4}W \times 32$ \\
				 \hline
				 \multicolumn{2}{c|}{ concat $\left[ \begin{array}{c} \textrm{conv2\_16, conv4\_3, branch\_1,} \\ \textrm{branch\_2, branch\_3, branch\_4} \end{array} \right] $} & $\frac{1}{4}H\times \frac{1}{4}W \times 320$ \\
				 \hline
				 fusion & $ \begin{array}{c} 3\times 3, 128 \\ 1\times 1, 32   \end{array} $ & $\frac{1}{4}H\times \frac{1}{4}W \times 32$ \\
				 \midrule
				 
				 \multicolumn{3}{c}{\textbf{Cost Volume}} \\
				 
				 \midrule
				 
				 \multicolumn{2}{c|}{Concat left and shifted right} & $\frac{1}{4}D\times \frac{1}{4}H\times \frac{1}{4}W \times 64$ \\ 
				 
				 \midrule
				 
				 \multicolumn{3}{c}{\textbf{Cost Aggregation}} \\
				 
				 \midrule
				 
				 3Dconv0\_x & $\left[ \begin{array}{c} 3\times 3\times 3, 32 \\ 3\times 3\times 3, 32  \end{array} \right] \times 2 $ & $\frac{1}{4}D\times \frac{1}{4}H\times \frac{1}{4}W \times 32$ \\ 
				 \hline
				 
				 3Dstack1\_1 & $\left[ \begin{array}{c} 3\times 3\times 3, 64 \\ 3\times 3\times 3, 64  \end{array} \right] $ & $\frac{1}{8}D\times \frac{1}{8}H\times \frac{1}{8}W \times 64$ \\ 
				 \hline
				 3Dstack1\_2 & $\left[ \begin{array}{c} 3\times 3\times 3, 64 \\ 3\times 3\times 3, 64  \end{array} \right] $ & $\frac{1}{16}D\times \frac{1}{16}H\times \frac{1}{16}W \times 64$ \\ 
				 \hline
				 3Dstack1\_3 & $\begin{array}{c} \textrm{deconv} \; 3\times 3\times 3, 64 \\ \textrm{add \textbf{3Dstack1\_1}}  \end{array} $ & $\frac{1}{8}D\times \frac{1}{8}H\times \frac{1}{8}W \times 64$ \\ 
				 \hline
				 3Dstack1\_4 & $\begin{array}{c} \textrm{deconv} \; 3\times 3\times 3, 32 \\ \textrm{add \textbf{3Dconv0\_2}}  \end{array} $ & $\frac{1}{4}D\times \frac{1}{4}H\times \frac{1}{4}W \times 32$ \\ 
				 \hline
				 
				 3Dstack2\_1 & $\left[ \begin{array}{c} 3\times 3\times 3, 64 \\ 3\times 3\times 3, 64 \\ \textrm{add \textbf{3Dstack1\_3}}   \end{array} \right] $ & $\frac{1}{8}D\times \frac{1}{8}H\times \frac{1}{8}W \times 64$ \\ 
				 \hline
				 3Dstack2\_2 & $\left[ \begin{array}{c} 3\times 3\times 3, 64 \\ 3\times 3\times 3, 64  \end{array} \right] $ & $\frac{1}{16}D\times \frac{1}{16}H\times \frac{1}{16}W \times 64$ \\ 
				 \hline
				 3Dstack2\_3 & $\begin{array}{c} \textrm{deconv} \; 3\times 3\times 3, 64 \\ \textrm{add \textbf{3Dstack1\_1}}  \end{array} $ & $\frac{1}{8}D\times \frac{1}{8}H\times \frac{1}{8}W \times 64$ \\ 
				 \hline
				 3Dstack2\_4 & $\begin{array}{c} \textrm{deconv} \; 3\times 3\times 3, 32 \\ \textrm{add \textbf{3Dconv0\_2}}  \end{array} $ & $\frac{1}{4}D\times \frac{1}{4}H\times \frac{1}{4}W \times 32$ \\ 
				 \hline
				 
				 3Dstack3\_1 & $\left[ \begin{array}{c} 3\times 3\times 3, 64 \\ 3\times 3\times 3, 64  \\ \textrm{add \textbf{3Dstack2\_3}}    \end{array} \right] $ & $\frac{1}{8}D\times \frac{1}{8}H\times \frac{1}{8}W \times 64$ \\ 
				 \hline
				 3Dstack3\_2 & $\left[ \begin{array}{c} 3\times 3\times 3, 64 \\ 3\times 3\times 3, 64  \end{array} \right] $ & $\frac{1}{16}D\times \frac{1}{16}H\times \frac{1}{16}W \times 64$ \\ 
				 \hline
				 3Dstack3\_3 & $\begin{array}{c} \textrm{deconv} \; 3\times 3\times 3, 64 \\ \textrm{add \textbf{3Dstack1\_1}}  \end{array} $ & $\frac{1}{8}D\times \frac{1}{8}H\times \frac{1}{8}W \times 64$ \\ 
				 \hline
				 3Dstack3\_4 & $\begin{array}{c} \textrm{deconv} \; 3\times 3\times 3, 32 \\ \textrm{add \textbf{3Dconv0\_2}}  \end{array} $ & $\frac{1}{4}D\times \frac{1}{4}H\times \frac{1}{4}W \times 32$ \\ 
				 \hline
				 
				 output\_1 & $\begin{array}{c} 3\times 3\times 3, 32 \\ 3\times 3\times 3, 1  \end{array} $ & $\frac{1}{4}D\times \frac{1}{4}H\times \frac{1}{4}W \times 1$ \\ 
				 \hline
				 
				 output\_2 & $\begin{array}{c} 3\times 3\times 3, 32 \\ 3\times 3\times 3, 1 \\ \textrm{add \textbf{output\_1}} \end{array} $ & $\frac{1}{4}D\times \frac{1}{4}H\times \frac{1}{4}W \times 1$ \\ 
				 \hline
				 output\_3 & $\begin{array}{c} 3\times 3\times 3, 32 \\ 3\times 3\times 3, 1 \\ \textrm{add \textbf{output\_2}} \end{array} $ & $\frac{1}{4}D\times \frac{1}{4}H\times \frac{1}{4}W \times 1$ \\ 
				 
				 \midrule
				 
				 \multicolumn{3}{c}{\textbf{For each output in [output\_1, output\_2, output\_3]}} \\
				 
				 \midrule
				 
				 upsampling & deconv $8\times 8\times 8$, stride=4 & $D \times H \times W$ \\
				 \hline
				 CENet & $\left[ \begin{array}{c} 3\times 3, 48 \\ 1\times 1, 1 \\ \textrm{sigmoid} \end{array} \right]$ & $H \times W$ \\ 
				 \hline
				 \multicolumn{2}{c|}{disparity regression} & $H \times W$ \\ 
				
				\bottomrule	
			\end{tabular}
		}
	\end{center}
	\label{network parameters}
\end{table}

{\small
	\bibliographystyle{aaai}
	\bibliography{egbib}
}
	
\end{document}